\documentclass[letterpaper, 10 pt, conference]{ieeeconf}  

\IEEEoverridecommandlockouts                              

\usepackage[noadjust]{cite}
\usepackage{caption}
\usepackage[hidelinks]{hyperref}
\usepackage{graphicx}
\usepackage{todonotes}

\title{\LARGE \bf
Non-invasive Cognitive-level Human Interfacing for the Robotic Restoration of Reaching \& Grasping}

\author{Ali Shafti$^{1,3}$ and A. Aldo Faisal$^{1,2,3,4}$
\thanks{*Research supported by eNHANCE (\href{http://www.enhance-motion.eu}{http://www.enhance-motion.eu}) under the European Union's Horizon 2020 research and innovation programme grant agreement No. 644000.}
\thanks{AS and AAF are with the Brain and Behaviour Lab, $^{1}$Dept. of Computing and $^{2}$Dept. of Bioengineering, $^{3}$Behaviour Analytics Lab, Data Science Institute, $^{4}$MRC London Institute of Medical Sciences, Imperial College London, SW7 2AZ, London, UK {\tt\small \{a.shafti,a.faisal\}@imperial.ac.uk}}%
}

\usepackage{atbegshi,picture}

\AtBeginShipoutNext{\AtBeginShipoutUpperLeft{%
  \put(\dimexpr\paperwidth-1cm\relax,-1.5cm){\makebox[0pt][r]{Accepted manuscript IEEE/EMBS Neural Engineering (NER) 2021}}%
}}

\begin{document}
\thispagestyle{empty}
\pagestyle{empty}
\maketitle
\begin{abstract}
Assistive and Wearable Robotics have the potential to support humans with different types of motor impairments to become independent and fulfil their activities of daily living successfully. The success of these robot systems, however, relies on the ability to meaningfully decode human action intentions and carry them out appropriately. Neural interfaces have been explored for use in such system with several successes, however, they tend to be invasive and require training periods in the order of months. We present a robotic system for human augmentation, capable of actuating the user's arm and fingers for them, effectively restoring the capability of reaching, grasping and manipulating objects; controlled solely through the user's eye movements. We combine wearable eye tracking, the visual context of the environment and the structural grammar of human actions to create a cognitive-level assistive robotic setup that enables the users in fulfilling activities of daily living, while conserving interpretability, and the agency of the user. The interface is worn, calibrated and ready to use within 5 minutes. Users learn to control and make successful use of the system with an additional 5 minutes of interaction. The system is tested with 5 healthy participants, showing an average success rate of $96.6\%$ on first attempt across 6 tasks. 
\end{abstract}


\section{Introduction}
\label{sec:introduction}
Assistive robotic systems represent an immense opportunity to overcome disabilities arising from motor impairments and allow those who have suffered from such losses to regain their physical interaction capabilities. From exoskeleton stroke rehabilitation devices \cite{bortole2015h2} to prosthetic systems \cite{farina2014extraction} assistive robotics present the possibility to enhance, augment, rehabilitate and/or replace human capabilities and functionalities, such as upper limb movement. 

\begin{figure}[tp]
    \centering
    \vspace{-20pt}
    \includegraphics[width=\columnwidth]{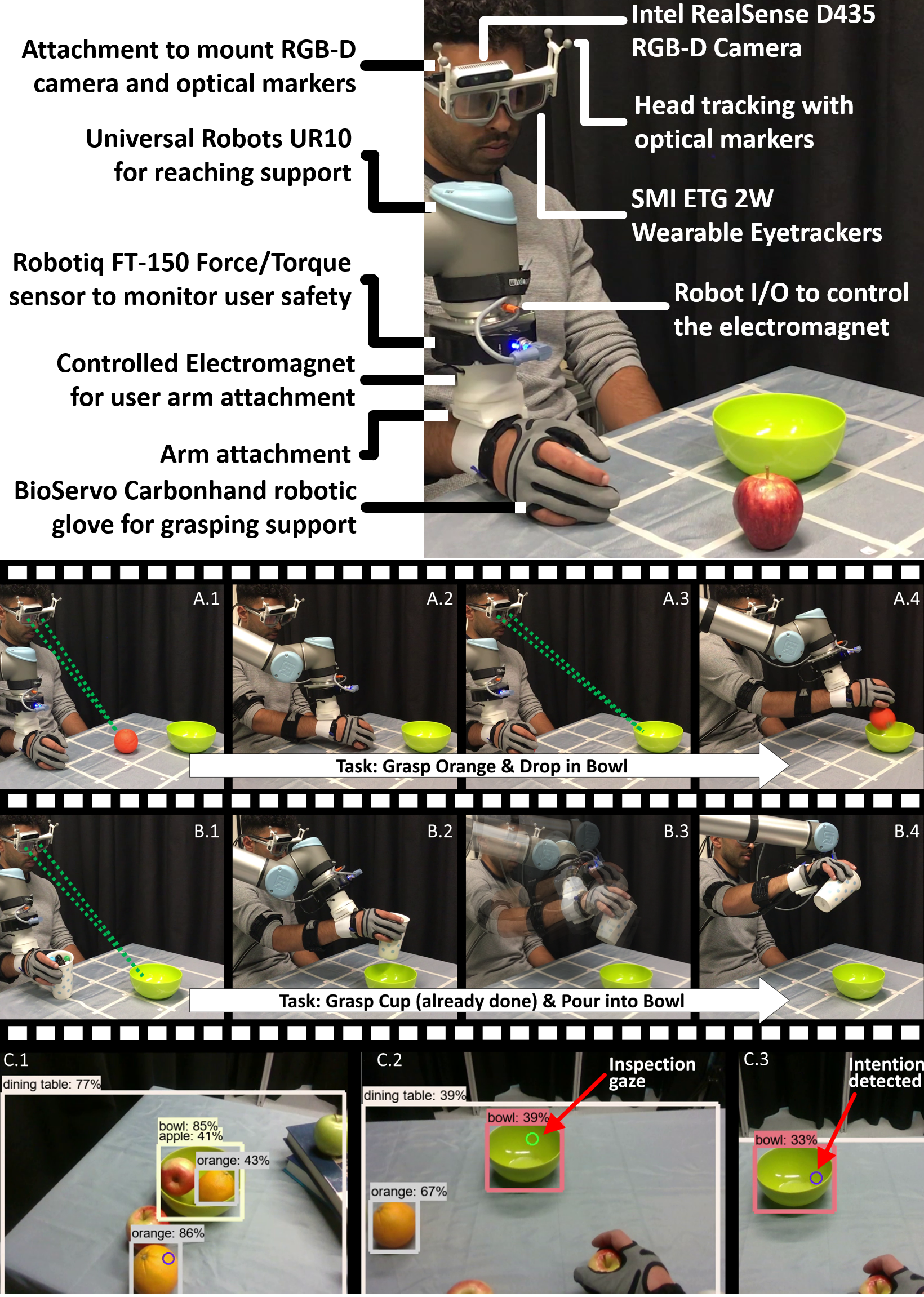}
    \caption{Our gaze-driven assistive robotic reach and support system. Top: The user wears eye-trackers that have been augmented with a depth camera for scene understanding and 3D gaze estimation. Head pose is tracked optically. The Universal Robots UR10 is used to actuate the user's arm to restore reach, and the BioServo Carbonhand robotic glove restores grasp. Bottom: \textbf{(A)} 1. User looks at the orange with an intention of action, this is detected by the system, task identified, motion planning performed, resulting in the robot arm moving the user's arm to the orange and the robot glove making the grip (2), once the orange is grasped, and their arm moved away to open the field of view, the user looks at the bowl with an intention of action, which results in the system assisting him in reaching the bowl and dropping the orange in it (4), \textbf{(B)} Here, the user has similarly grasped a cup, which, once the user looks at the bowl with an intention of action, changes the rest of the sequence to one of pouring, in (2) we see reach to bowl, in (3) and (4) changing of pose to enable pouring. \textbf{(C)} Examples of the egocentric view of the user, with gaze, object labelling and intention decoding running.}
    \label{setup_page1}
\vspace{-5pt}
\end{figure}

The effective use of robotic systems as motor assistive devices relies heavily on safe and efficient human-robot interfaces that allow the reliable decoding of human intention of action. Neural interfaces have been thoroughly explored within this context. The use of surface EMG is one non-invasive approach to this, and has been widely applied in control of robotic prostheses \cite{farina2014extraction}, though in some cases involving invasive surgical nerve transfer procedures to create additional recording sites for myoelectric control signals \cite{hargrove2017myoelectric}. Other approaches involve Brain-Machine Interfaces. Musallam et. al. have shown the possibility of identifying high-level cognitive signals, through implanted electrode arrays in monkeys, that define movement goals of reach in visual coordinates \cite{musallam2004cognitive}. Hochberg et. al. show the use of cortical neuronal ensemble signals obtained through a surgically implanted microelectrode array for the control a robotic arm in reach and grasp tasks by two tetraplegic users \cite{hochberg2012reach}. Collinger et. al. implant two 96-channel intracortical microelectrodes on an individual with tetraplegia and show that after 13 weeks of training, they can control 7 degrees of freedom with an athropomorphic robotic arm to fulfil tasks \cite{collinger2013high}. Ajiboye et. al. use a similar implant approach with a tetraplegic participant, allowing them to reach and grasp with their own limbs through the combination of robotic support and functional electrical stimulation \cite{ajiboye2017restoration}. While these results show the big potential and impact of robotic systems as assistive devices, they all rely on invasive, surgical procedures followed by long training times.

Here, we present the use of a non-invasive and intuitive robot interface: our eyes. Eye-movement is often preserved in the presence of other severe motor impairments \cite{abbott2012ultra}, and can serve as a window to our intentions: research has shown that our gaze patterns change depending on the task at hand \cite{borji2014defending}. Finally, eye movements as the natural interface between humans and their surrounding environment, make for an intuitive control interface. We previously showed the feasibility of using the eyes as a means to control robotic systems, e.g. to augment human capability \cite{noronha2017wink}, allowing to draw and write while eating and drinking \cite{dziemian2016gaze}, as well as assisting them in reaching and grasping using their own arm and hand \cite{maimon2017towards,shafti2019gaze}. We have also demonstrated other multi-modal approaches to head-gaze estimation \cite{sim2013head} and neural interfaces \cite{fara2013robust} for BMI use. Others have shown the feasibility of using gaze as an interface for robot control, be it for augmentation in cases such as surgical robotics \cite{mylonas2006gaze}, or assistive robots that fulfil tasks for the user \cite{aronson2020eye}. Here, we present an expansion of our previous work \cite{shafti2019gaze}, enabling users to reach and grasp using their own arm and hand, assisted by a robotic arm and glove, with gaze as the interface. 


\section{Materials and Methods}
\label{sec:methods}
\begin{figure}[tp]
    \vspace{4pt}
    \centering
    \includegraphics[width=\columnwidth]{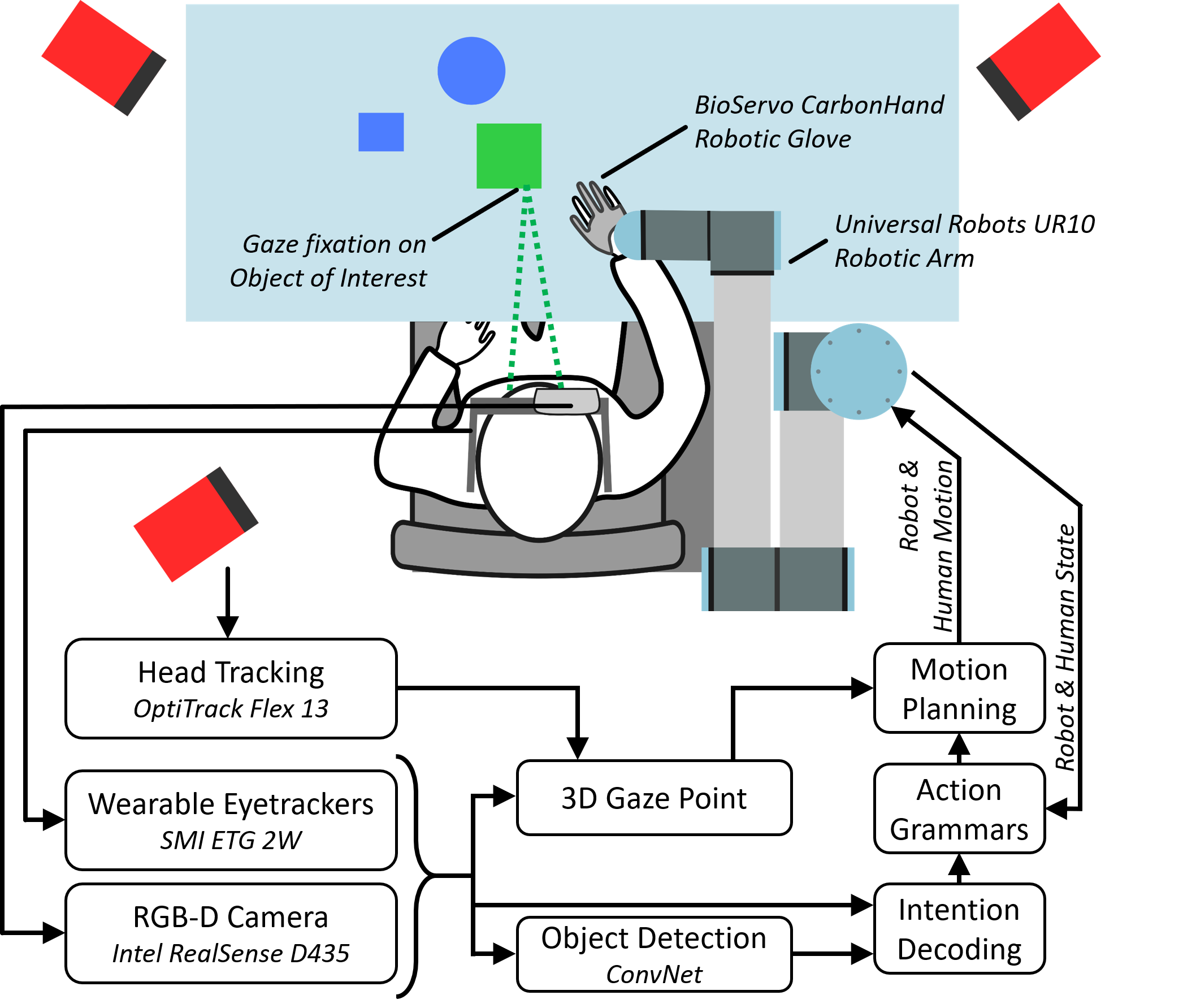}
    \caption{Block diagram of our system architecture.}
    \label{methods_blockdiag}
\vspace{-5pt}
\end{figure}

The overview of our system's architecture can be see in Figure \ref{methods_blockdiag}. We use the SMI ETG 2W eye trackers (SensoMotoric Instruments Gesellschaft f\"{u}r innovative Sensorik mbH, Teltow, Germany), augmented with the Intel RealSense D435 RGB-D camera (Intel Corporation, Santa Clara, California, USA) to provide the 3D context of gaze and surrounding environment. We use Optitrack Flex 13 cameras (NaturalPoint, Inc. DBA OptiTrack, Corvallis, Oregon, USA) to track the user's head pose through optical markers (see Figure \ref{setup_page1}). We use our own method for absolute 3D gaze point estimation which we presented in detail in \cite{shafti2019gaze}.

To understand the context of gaze, we create a real-time object detection system using a Convolutional Neural Network (CNN) implemented in TensorFlow, which runs on the RGB stream and labels objects in the user's egocentric field of view, details previously presented in \cite{auepanwiriyakul2018semantic}. Through this we can infer the intention of the user. Intention of action is identified based on the user's gaze relative to object bounding boxes: we assign the right-most $1/3$ of the bounding box to indicate intention through fixation. This is applicable to all objects, and gives our users the agency to inspect and indictate intent when required.

Once an action intention is detected, our system identifies the sequence of action for the user's intended task by considering the current human-robot state and parsing it through our Action Grammars approach. Our behaviour, similar to our language, can be represented as having rules (grammar) on how to combine actions and in which order to use them to create a meaningful sequence. By extracting the grammars of action by observing human behaviour, we are able to reduce the dimensionality of the robot action selection problem, as the system will be facing a smaller set of action choices based on the current human-robot state. For the purposes of this work, we consider a simulated dining table scenario, including small and large containers, and non-container objects, leading to possibilities of pick and place, as well as pick, pour (from small container to large container) and place. Grammars for these sequences are hand-derived, and are parsed through a finite state machine implementation. Further details on this implementation can be seen in \cite{shafti2019gaze}.

We use the Robot Operating System (ROS) \cite{quigley2009ros} for integration, running on a Linux workstation (ROS Kinetic, Ubuntu 16.04). The Motive software used for optical tracking as well as the eye trackers integrated with the RGB-D camera run on Windows 10 workstations networked with the ROS master. Our robotic setup consists of the Universal Robots UR10 (Universal Robots A/S, Odense, Denmark) and the soft-robotic BioServo Carbonhand (Bioservo Technologies AB, Kista, Sweden). Both robot systems are linked to the human and controlled through ROS -- the UR10 for reaching, and the Carbonhand for grasping purposes. Users of the system will wear the Carbonhand, and attach their arm to the UR10's end-effector at their wrist. We use a fixed-joint arm attachment to have full control over the user's hand pose for improved manipulation. This, however, means that we need to motion plan for human arm kinematics. We calculate the optimal reaching orientation for the grasp of each object, based on the object’s location, with respect to the user’s elbow location. The default start point is selected to keep the user’s lower-arm parallel to the ground, and the upper-arm perpendicular to it. Thus, by knowing the initial location of the user’s elbow, and the target point, we are able to identify the necessary waypoints for the robotic arm to reach, with an optimal orientation of the arm. These are selected by initially keeping the elbow location constant, and positioning the forearm so that it is aligned with the line connecting the elbow and the target location in space. From there, the arm can be safely moved over the elbow-target line to approach the target location. Once reaching is complete, the robotic glove can be activated to grasp or drop. We combine the glove's sensor and motor data to infer grasp success, which informs our action grammars module for further planning\footnote{A video of the system can be seen here: \href{https://youtu.be/VHcipV4UAUs}{https://youtu.be/VHcipV4UAUs}}. 

For added safety, our fixed attachment uses an electromagnet connected through a Robotiq FT-150 force/torque sensor (Robotiq Inc., Quebec, Canada) to the UR10’s end-effector. This allows us to have closed-loop control on the release of the user’s arm for safety, i.e. in cases where unsafe forces or torques are detected at the attachment (see Figure \ref{setup_page1}).


\section{Experimental Evaluation}
\label{sec:experiments}
Our experiment involves simulated dining table scenario tasks. The objects in use are: apple, orange, cup, bowl and dining table. The tasks can then be defined as follows:

\begin{itemize}
    \item Grasp \{Orange, Apple\}, drop in/on \{Bowl, Table\}.
    \item Grasp \{Cup\}, pour into \{Bowl\}, drop on \{Table\}.
    \item Grasp \{Cup\}, drop on \{Table\}.
\end{itemize}

We tested all above combinations, leading to a total of 6 options, with 5 participants. Participants are allowed to repeat each option up to 5 times in case of failure, but if succeeded, move immediately to the next option. Each participant starts by first receiving a description of the entire setup and what to expect from the experiment. They then wear the robotic glove and have their arm attached to the robotic arm. The eye-trackers are worn and participants go through the initial calibration. Once calibration is complete, the participants are given 5 minutes to test out the setup and get used to the environment. Trials are randomised in terms of task and object placement. Performance in completing tasks, and subjective evaluation of the system’s usability through the System Usability Scale (SUS) questionnaire \cite{brooke1996sus} are reported.


\section{Results \& Discussion}
\label{sec:results}
We tested the system with 5 healthy participants who were directed to fully relax their right arm and hand and not to help the system with actuation, to simulate paralysis.

\subsection{Task Performance}
Across all participants, and for all tasks, for the reach actions, for all tasks, $100\%$ of participants were successful on their first attempt. A successful reach would mean computing the 3D gaze point of interest for the user, detecting an intention of action, motion planning and moving their arm to the target point, with the correct pose to enable the grasping of the intended object. For the grasping action, i.e. grasping the intended object after a successful reach, $100\%$ of participants made successful grasps on their first attempt. For the subsequent dropping or pouring of objects, except for a single case where participant 2 had a failed pour and only succeeded on their second attempt, all other attempts across all participants and all tasks were limited to 1. Therefore, $96.6\%$ of drop/pour actions were successful on the first attempt. For drop actions alone this was $100\%$. Therefore, at task level, except for the one pouring failure, all participants succeeded at all 6 tasks on their 1st attempt.

These results show the feasibility of using the combination of gaze and action grammars as an interface for assistive robotic systems. All 5 participants, who were previously naive to our non-invasive system, were capable to use it successfully after an initial 5 minute calibration, followed by 5 minutes of free use for training.

\subsection{Subjective questionnaire}
\begin{figure}[tp]
    \centering
    \vspace{4pt}
    \includegraphics[width=\columnwidth]{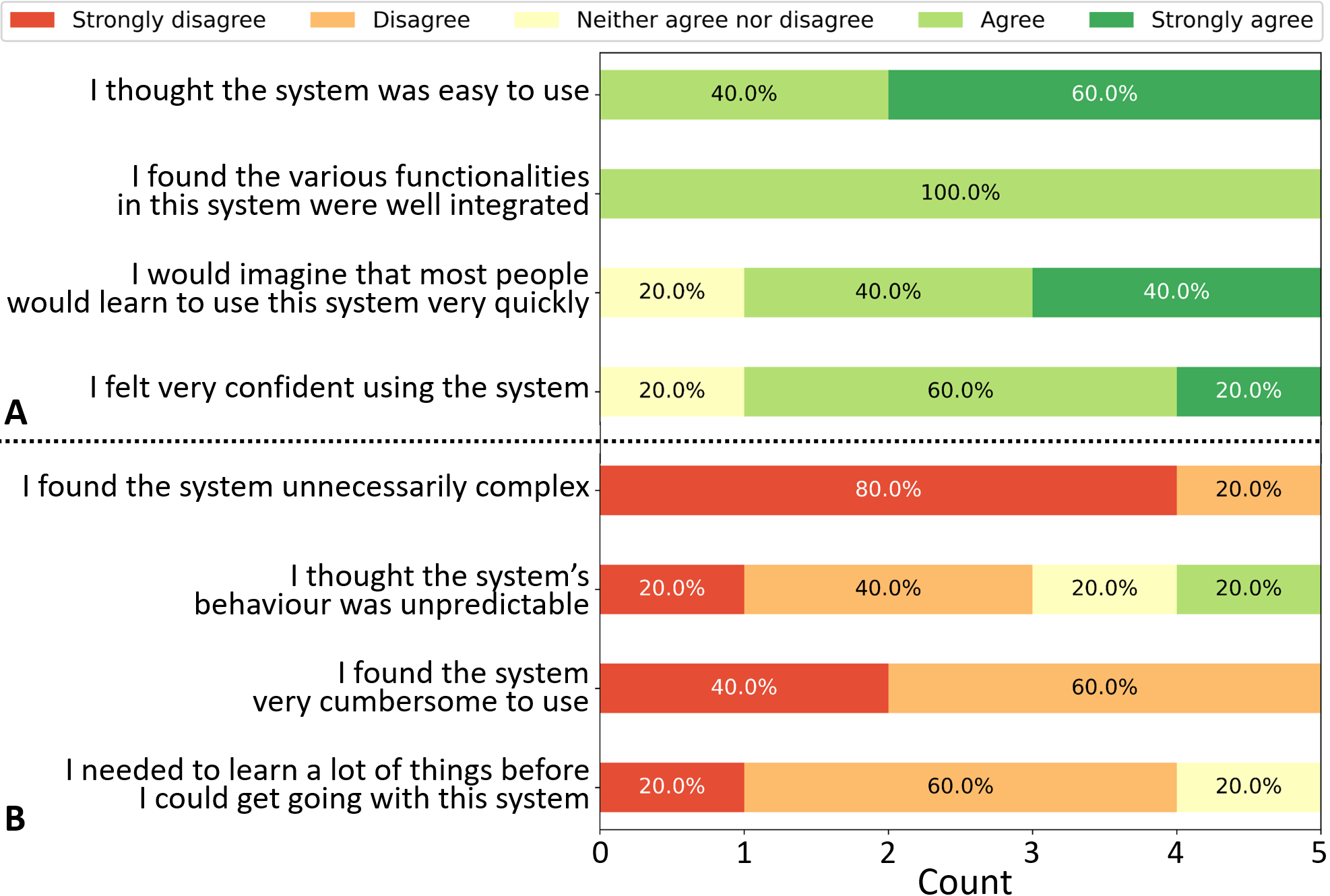}
    \caption{Results of the System Usability Scale (SUS) questionnaire filled by our 5 participants after their experience with our system \textbf{(A)} positive statements, Agreeing is good, \textbf{(B)} negative statements Disagreeing is good.}
    \label{results_sus}
\vspace{-5pt}
\end{figure}

All participants filled a System Usability Scale \cite{brooke1996sus} form after the experiment. Results are shown in Figure \ref{results_sus}, separated by positive and negative statements respectively, i.e. Agreeing is the good result in Figure \ref{results_sus}.a whereas Disagreeing is the good result in Figure \ref{results_sus}.b.

For all of the positive statements shown in Figure \ref{results_sus}.A we see no participant disagreeing. All 5 participants agree that they found the system easy to use, with $60\%$ strongly agreeing. All participants found that the various functionalities of the system were well integrated. On whether they imagine the system would be easy to use for most people, $80\%$ agree (half of them `strongly'), whereas $20\%$, i.e. 1 participant, is neutral. On whether they felt confident when using the system, $80\%$ agree and 1 participant is neutral. 

Looking at the negative statements summarised in Figure \ref{results_sus}.B, we see that $100\%$ of participants disagree ($80\%$ strongly disagree) with the system being characterised as unnecessarily complex. On whether the system's behaviour was unpredictable $60\%$ diagree, $20\%$ are neutral and $20\%$ agree (i.e. 1 participant). On whether the system is cumbersome to use, $100\%$ of participants disagree, $40\%$ strongly. On whether participants needed to learn a lot to use the system, we see $80\%$ disagreeing, and 1 participant neutral. 

Overall, we see a clear majority agreeing on positive statements, and disagreeing on negative ones, indicating that our participants are happy with the system. Note that due to the short calibration and learning requirements of our setup, the entire experiment from entering the room until departure, took a maximum of 30mins for all participants. This is perhaps not enough time to get fully used to the system. While majority agree that they felt confident with the system, we believe this would have improved with more time using the system. On the system's behaviour, we have one participant characterising the system's behaviour as unpredictable. This is perhaps the main message: need for further feedback mechanisms, so that participants are better aware of the course of action by the robotic system. Our Action Grammar approach allows us to explain the behaviour of the robot on a human-interpretable level, i.e. task/action level; feedback mechanisms will be pursued as future work.


\section{Conclusions} 
\label{sec:conclusions}
We present a cognitive-level brain interface through gaze for assistive robotic restoration of reaching and grasping. While many successful efforts within Brain-Machine Interfaces research for assistive robot control is based on invasive, surgically implanted and training-heavy methods \cite{musallam2004cognitive,hochberg2012reach,collinger2013high,ajiboye2017restoration}, our approach to interfacing the human mind and decoding their intentions is non-invasive, requires 5 minutes to be worn and calibrated, and shows a very high success rate after 5 minutes of training in healthy controls. Musallam et. al. argue for the use of cognitive level signals representative of goals in visual coordinates \cite{musallam2004cognitive}. Here we take this concept further by removing the need for any invasive procedures, and interfacing the human entirely through their eye-movements and on the cognitive level. 

Our system detaches users from the multiple degrees of freedom involved in the robot control and motion planning problem, allowing them to simply look at an object that they want to manipulate, with an intention to do so, at which point the system will handle all the lower-level controllers necessary to perform their intended task successfully, safely and efficiently. Our very positive subjective questionnaire results support this, showing that participants find the system easy to use/learn, well-integrated and felt confident using it, among other positive feedback. Based on the feedback we have received, we are improving our system with additional features, such as 3D mapping of the task environment, and a dedicated kinematic solver for the human arm, enabling the system to to plan for and avoid obstacles, but also to move the human arm in natural human motion paths to improve the user experience while conserving safety and interpretabiltiy.

\addtolength{\textheight}{-12cm}   








\bibliographystyle{ieeetr}
\bibliography{references}

\end{document}